# Tumor Classification and Segmentation of MR Brain Images

Tanvi Gupta[a,*], Pranay Manocha[b], Tapan K. Gandhi[a], R.K. Gupta[c], B.K. Panigrahi[a]

[a]Department of Electrical Engineering, Indian Institute of Technology, Delhi, India, 110016
[b]Department of Electrical Engineering, Indian Institute of Technology, Guwahati, India, 781039
[c]Fortis memorial research institute, Gurgaon, India,122002

## Abstract

The diagnosis and segmentation of tumors using any medical diagnostic tool can be challenging due to the varying nature of this pathology. Magnetic Resonance Imaging (MRI) is an established diagnostic tool for various diseases and disorders, and plays a major role in clinical neuro-diagnosis. Supplementing this technique with automated classification and segmentation tools is gaining importance, to reduce errors and time needed to make a conclusive diagnosis. In this paper a simple three step algorithm is proposed; (1) identification of patients that present with tumors, (2) automatic selection of abnormal slices of the patients, and (3) segmentation and detection of tumor. Features were extracted by using discrete wavelet transform on the normalized images, and classified by support vector machine (for step (1)) and random forest (for step (2)). The 400 subjects were divided in a 3:1 ratio between training and test with no overlap. This study is novel in terms of use of data, as it employed the entire T2 weighted slices as a single image for classification and a unique combination of contralateral approach with patch thresholding for segmentation, which does not require a training set or a template as is used by most segmentation studies. Using the proposed method, the tumors were segmented accurately with a classification accuracy of 95% with 100% specificity and 90% sensitivity.

*Corresponding author
Email address: tan0036@gmail.com (Tanvi Gupta)

1. Introduction

The human brain is formidably complex entailing a host of factors such as age, gender, ethnicity and personal medical history. Diagnosis of brain abnormalities such as degenerative, infectious, ischemic or malignant are done using the Magnetic Resonance Imaging (MRI) which is an effective standardized neuro-imaging tool. A routine brain imaging protocol includes T1-weighted, T2-weighted, Fluid attenuated inversion recovery (FLAIR), Gadolinium-enhanced T1-weighted images. The mode of data acquisition is gradually shifting from two-dimensional to three-dimensional imaging. This results in a large volume of data per patient, for which the analysis is both time consuming and prone to error. This makes computer-aided detection desirable as an aide to the radiologist.

Tumors are atypical cells multiplying out of control. These may vary in size, location and type. They show a spectrum of atypia from benign to malignant. It is usually variegated with high grade and low grade tumor cells, necrosis and edema. Therefore, it is daunting to train a computational system to identify and segment the region of interest, making it the pathology of interest.

Various methods are used for automated disease classification and tumor segmentation, each with their own restrictions. The pathologies that are more commonly studied for classification purposes are degenarative diseases like Alzheimer's [1, 2, 3, 4], Parkinson's and Schizophrenia [5]. These affect the entire brain posing less of a challenge as far as classification is concerned as the effected brain varies significantly from physiology. The studies are primarily limited by a small dataset [6, 7, 8, 9, 10] mostly taken from medical libraries that are available on the internet, like the Harvard School Medical Library [11, 12, 13, 14, 15]. Moreover, the data sets often lack header information. Mostly, the images used for classification and segmentation are taken as one slice per patient from a two dimensional scan set [16, 10, 15]. However, the lesion usually does not appear only in one slice and hence limits the training and test data to certain slices of the brain, making it difficult to use in a clinical scenario.

Methods like Multi-Geometric Analysis (MGA) [15] and entropy based features using discrete wavelet transform [14], have been used for feature extraction. Detailed extraction methodology and normalization techique used for images is not discussed in most of the studies. Principal Component Analysis (PCA) is a commonly used unsupervised learning method for feature selection and reduction which increases the computation time and can eliminate certain data that is essential for classification. Most studies use PCA without justification or mention of the computational time [15, 10, 17, 18]. Many classification techniques such as support vector machine (SVM) [15, 1, 3, 13, 19, 2, 20, 21, 4], Artificial neural network (ANN) [16, 11, 17], Probablistic neural network (PNN) [22], Linear discriminant analysis (LDA) [10], k-Nearest neighbour (k-NN) [17] have been employed based on their own merits. Template comparison to a normal atlas of the human brain has also been used for classification [23], which would be limited to certain data and may not be able to correctly classify variants from normal physiology. The patient data under analysis is linear in nature and therefore, linear classifiers like SVM and random forest are useful. The success of the classification techniques is dependent on the feature set available and pre-processing of images used.

Convolutional neural network (CNN) [24, 25, 26], k-NN, Fuzzy networks [8] and other feature based techniques are used for segmentation which requires a large data base usually acquired from Multimodal Brain Tumor Image Segmentation Benchmark (BRATS) 2013. Methods like fuzzy c-means algorithm produce multiple segmented images depending on the number of clusters chosen and thus the final segmented image must be manually selected. Other methods use a normal template which may vary in intensity, could be a normal variant and may not be very accurate for tumor segmentation purposes.

The motivation of this study was to combine classification and segmentation techniques to form a robust system for identification of tumors using all the slices of each patient as a single component to classify patients into two classes of pathology and physiology. A multivariate data set comprising T2-weighted and FLAIR images from 400 patients, classified by the radiologist into normal and abnormal (etiology being tumor) has been considered. Tumors exist in multiple locations and vary between Glioblastomamultiforme (GBMs) and Low Grade Gliomas (LGGs). T2 and FLAIR sequences for these patients have been processed to classify and segment tumor. No feature selection method is employed and the algorithm is executed with a short computation time of 375 secs that includes training, testing and segmentation of all patients in question. It should be noted that the total number of images being processed are roughly 300 images for 3D data and 30 images for a 2D transverse data. Since 12 slices are used for classification and segmentation, after pre-processing of the data, the total images are 4800 (12 x 400). The patients that were identified to have tumors were further filtered to select only the slices with tumor, upon which tumor segmentation was performed. The segmentation is done without the use of a training set or a normal template.

This paper is structured as follows: Section 2 gives the step by step method followed for classification and segmentation. Section 3 analyses and discusses the results obtained and Section 4 concludes the work done.

2. Methodology

This work has been divided into three sections to systemically segment tumors. The first section filters patients with tumors (group1) in contrast to patients without tumors (group2), the second further filters group1 into the slices with tumors, and the third segments the tumor and stores the 3D data with the marked region of interest. The process flow for the proposed methodology is shown in figure 1.

2.1. Data used

Data was collected from a single setup with all imaging performed on a 3-Tesla MR scanner (Philips Healthcare, Netherlands) under the expert super-vision of a neuroradiologist (R.K.G.) with more than 30 years of experience.

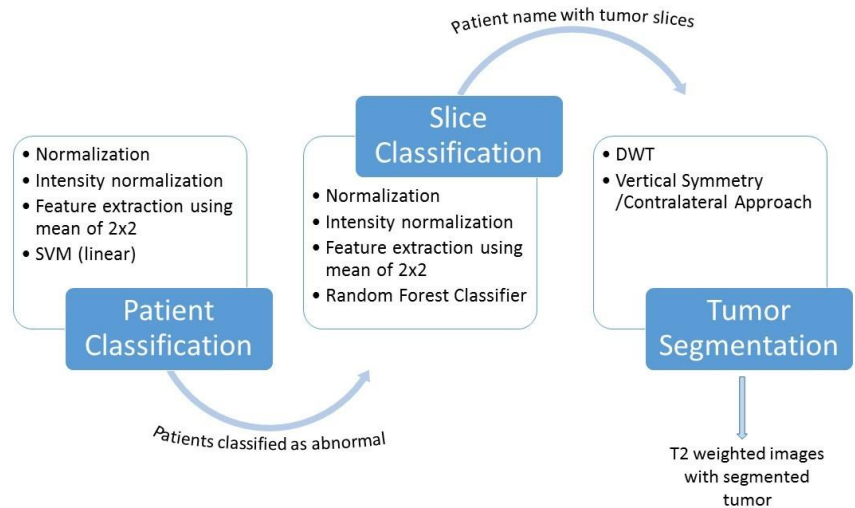

Figure 1: Process for classification and tumor segmentation.

Test subjects were classified into normal, referred to as group2 (n=200) and with findings, referred to as group1 (n=200). Training and testing groups were randomly selected from each group in the ratio of 3:1, respectively and without duplication.

Age of group1 ranged from 2 to 82 years with a mean of 43.64 ± 18.41 years; male:female ratio of 1.82. Among these, the training group had a male:female ratio of 1.68 with a mean age of 32.66 ± 18.01 and test group had a male:female ratio of 2.33 with a mean age of 43.92 ± 21.44. Age of group2 ranged from 4 to 71 years with a mean of 35.36 ± 14.44 years; male:female ratio was 1.27. Among these, the training group had a male:female ratio of 1.17 with a mean age of 27.01 ± 14.71 and test group had a male:female ratio of 1.63 with a mean age of 33.42 ± 16. Tumors varied in size, shape, location, tumor grade and multiplicity.

Volumetric T2-FLAIR, and volumetric T2-weighted images (or 2D) were taken. T2-weighted MR images use long TE (echo time) and TR (relaxation time) to resolve water from fat; the former has longer time span for both. FLAIR is a T2-based pulse sequence which nullifies fluid and displays the pathology prominently [27].

All images were normalized to 64 x 64 matrix of 10 mm slice thickness yielding a homogeneous element of 16 slices.

2.2. Image Post-processing

For all the three steps the image post-processing steps followed were the same and are described in this section.

2.2.1. Normalization

The patient data, either volumetric or 2-dimensional, was normalized using SPM8 toolbox in MATLAB using the respective T2.nii template. The size taken for normalization was [2,2,10] which indicates that the image size for each slice is 64x64 with 10 mm slice gap, thus generating a total of 16 slices per patient. The volumetric normalization was done as per the equations 1 to 4. Volumetric normalization maximizes the overlap of voxels of the images being processed X and the template $X_0$ as seen in equation 1, where T is a rigid body transformation.

$$X_0 = \{X_0 : X_0 \in X \cap T(X^0)\} \quad (1)$$

If $F(X_0)$ is the set of intensities for the overlapped voxels in $X_0$ with mean $\bar{f}$ and $G(X_0)$ is the set of intensities for the overlapped voxels of X with mean $\bar{g}$ the normalized correlation coefficient (NCC) is given in equation 2.

$$NCC(F, G) = \frac{1}{N_0^2} \frac{\Sigma_{x_0 \in X_0}(f(x_0) - \bar{f})(g(x_0) - \bar{g})}{\sigma_f \sigma_g} \quad (2)$$

$$\sigma_f = \frac{1}{N_0} \sqrt{\Sigma_{x_0 \in X_0}(f(x_0) - \bar{f})^2} \quad (3)$$

$$\sigma_g = \frac{1}{N_0} \sqrt{\Sigma_{x_0 \in X_0}(g(x_0) - \bar{g})^2} \quad (4)$$

The first and last 2 slices are discarded as they contain very little useful information. Intensity normalization was performed for each slice of each patient, where the intensity range is changed to 0-1 by dividing the value of each pixel by the highest pixel value of the slice as seen in equation 5 where x is the pixel value and i and j represent row and column of the image respectively. Then the data was compiled to yield a homogeneous element of 12 slices for further processing and classification.

$$x_{i,j} = \frac{x_{i,j}}{\max(X)} \quad (5)$$

2.2.2. Feature Extraction

Each slice was divided by a 2x2 grid and the mean of the voxel values in the grid was considered as the feature of the grid as shown in equation 6. Therefore, 32x32 (64x64/2x2) features were obtained per slice and 32x32x12 features were obtained for each patient.

$$x_{i,j} = \frac{x_{i,j} + x_{i+1,j} + x_{i,j+1} + x_{i+1,j+1}}{4} \quad (6)$$

2.3. Patient classification by Support Vector Machine

Support Vector Machines (SVMs) are based on the concept of decision planes that define boundaries between different classes of objects. A decision plane is one that separates a set of objects having different class memberships. There are different types of decision planes like Linear, Quadratic and Polynomial which fit data into different classes for classification [28, 2, 29, 30, 31, 32, 33, 34]. For the linear kernel, the equations for solving the hyper plane equations are as given in equations 7 and 8.

$$w^T + b = +1 \text{ for } d = +1 \quad (7)$$

$$w^T + b = -1 \text{ for } d = -1 \quad (8)$$

Where, w is a weight vector, x is the input vector and b is a bias. d is the margin of separation between the hyper plane and the closest data point for a given weight w and bias b. Optimal decision plane is the one which maximizes the margin of separation d.

## 2.4. Slice Identification

Group 1 patients were processed at this stage to identify the slices of interest with each slice considered as a separate feature set. The number of rows are equal to the product of number of patients and slices (12) and columns are equal to the square of 32, for the resulting feature matrix.

### 2.4.1. Random Forest Classifier

Random forests is a learning method for classification, regression and other tasks. It operates by constructing a multitude of decision trees for training and gives the class as the output. It is a combination of multiple classification or regression trees combined to improve the accuracy of training and therefore classification [35, 36, 37]. The slices that are used differ from each other significantly thus linear SVM was insufficient for this classification. Random forest with 25 trees was used to obtain a high sensitivity.

## 2.5. Tumor Segmentation

With the abnormal slices known for each patient the data is processed for tumor segmentation. The data used is the normalized T2 data which has undergone intensity normalization and has been re-sized to 64 x 64 for ease of processing.

### 2.5.1. Discrete Wavelet Transform (DWT)

DWT helps visualize images in the time-frequency domain using low and high pass filters to decompose it. The function in its discrete form is given by the following equations:

$$cA_{j,k}(n) = \sum_{n}^{h_i} f(n) l_j^*(n - 2^j k) \qquad (9)$$

$$cD_{j,k}(n) = \sum_{n}^{h_i} f(n) h_j^*(n - 2^j k) \qquad (10)$$

$l(n)$ and $h(n)$ are the low and high pass filters, respectively, and $cA_{j,k}$ and $cD_{j,k}$ represent the approximation components containing the low frequency information and detailed components containing the high frequency details of the image which basically are the edge of the image k-space [38, 39, 40, 41, 14]. Tumor segmentation requires an approximation component image as a base contrast of the tumor. Eliminating the high frequency components removes the edges including skull patterns which might cause errors in segmentation. The approximation image is re-sized to 64x64 and used for tumor segmentation.

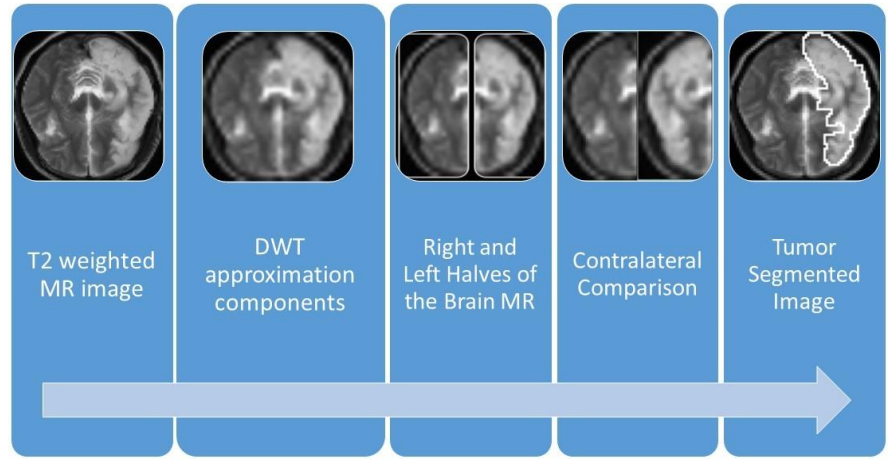

Figure 2: Flow chart for tumor segmentation.

2.5.2. Thresholding by Contralateral Comparison

The Vertical Symmetry or the Contralateral approach is dependent on the fact that the bilateral cerebral hemispheres are comparable. The presence of a tumor distorts the symmetry of the brain, and hence this method is appropriate. Only the slices with tumors as classified by the Random Forest classifier are taken and analyzed further for tumor segmentation. The steps for segmentation are as follows and are shown in figure 2:

1. If any slice demonstrates tumor, the neighboring 3 slices are analyzed to locate tumor margins and hence verify the prediction.

2. Next all stray tumor slices are removed i.e those slices with no adjoining slices, and therefore, removing the confounding features.
3. Then the remaining slices are made continuous i.e. for example if slices 6-7 and 11-12 were found to qualify as above, then all slices from 6 to 12 are considered for segmentation.
4. To segment the tumor (voxels identified as tv), the right and left half of the brain are compared to find the points of intensity difference above a threshold as shown in equation 11. This is advantageous as the contralateral side serves as the control and training set is not required.

$$tv_{i,j} = x_{i,j+n/2} > x_{i,n/2-j} + \text{threshold} \qquad (11)$$

where n is the total number of columns and $j=0,1,2...n/2$.

5. A 4x4 section is created around the selected points and if, the number of points in the patch is less than a threshold, the patch is removed. This is done assuming that pathology shall be larger than the patch size considered, and our methodology removes smaller asymmetries which qualify for normal variants.
6. The remaining sections that are considered to be tumors are delineated.

## 3. Results
### 3.1. Patient Classification

In the first step, all patients were classified into group1 and group2 using SVM. The accuracy, sensitivity and specificity were calculated as per equations 12, 13 and 14 [17, 11, 42, 43]. Testing with T2WI and FLAIR yielded accuracy, sensitivity and specificity of 92.00%, 90.00%, 94.00% and 88.78%, 84.91%, 92.60%, respectively.

Group2 patients were subjected to the next step. The algorithm was run on MATLAB R2014a on a 4GB RAM, 2.3GHz i5 processor and took 97 seconds to process the algorithm.

$$\text{Accuracy} = \frac{\text{Correctly classified data}}{\text{Total data}} \times 100\% \qquad (12)$$

$$\text{Sensitivity} = \frac{\text{Correctly identified abnormal data}}{\text{Total abnormal data}} \times 100\% \qquad (13)$$

$$\text{Specificity} = \frac{\text{Correctly identified normal data}}{\text{Total normal data}} \times 100\% \qquad (14)$$

3.2. Slice Selection

Testing with T2WI yielded sensitivity of 77.52%. The algorithm outed the slice numbers that are abnormal. This result acts as the input to the next stage.

3.3. Tumor Segmentation

These slices are thereafter, presented to the next step where they undergo segmentation and the output is a nifti file of the T2 weighted images which include a white line which demarcates the tumor region, as exemplified in figure 3. The false positive from the first stage do not have any tumor regions segmented, thus increasing the specificity of the overall algorithm to 100% and accuracy to 95%. Figure 3 shows an example of the corresponding slice for a normal patient, patient with tumor and the post-segmentation image of the respective patient.

3.4. Discussion

Automated classification of abnormal images and tumor segmentation is no longer a pre-clinical research tool. Our algorithm demonstrates that it can be applied to clinical data. To be applicable to clinical data all slices must be included for classification with a standardized normalization procedure. As compared to work done in previous literature, in the present approach, a combination of classification and tumor segmentation is used to achieve a higher accuracy in terms of identifying patients with tumors and identifying the slice and area in which the tumor occurs [18, 44, 22, 45, 46]. A multi-variate data set of 400 patients (T2-weighted images) are used where all the images taken in the scan are considered unlike most studies which are limited by data [18, 22, 44, 45, 46, 47, 12] and use only one image per patient [18, 22]. Each patient set is treated as an image thus increasing accuracy and reliability. If only one slice is given as training, it implies that a diagnostician would be required to supply the tumor image for classification, thus defeating the classification purpose. Here the entire data taken from a scan is fed into the program to give a result of pathology or physiology as well as slice and area in which the tumor appears. The data is taken from a clinical setup and the parameters of the scanner and protocols are known. The images are normalized and then intensity normalization is done to ensure that all images are scaled. The details of normalization, sequence proto-col used, details of feature extraction and size of feature matrix are limited in past studies [48, 47, 12].

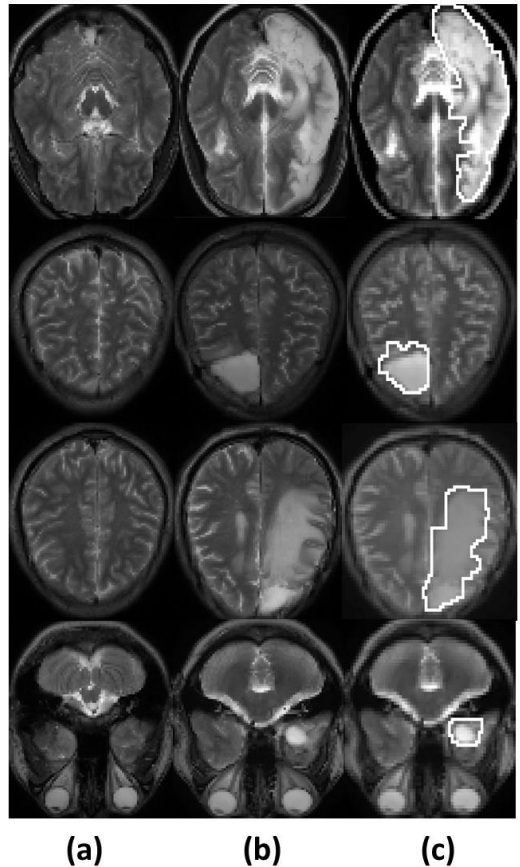

Figure 3: Tumor segmentation: (a) Slice of normal patient, (b) Slice of patient with tumor, (c) Slice with segmented tumor.

A linear SVM classifier was used for patient classification in the first step, as the features when plotted appeared linear. This is a basic classifier which reduces computational time as it is used without a feature reduction method namely PCA. With a large feature matrix, computation time for covariance matrix and eigen vectors is high. Therefore the time increases on use of PCA. Most studies that use PCA do not justify the use of PCA in terms of accuracy or feature reduction and do not report computation time [18]. We tested the algorithm with and without PCA and found that there was no significant change in accuracy whereas the computation time increased significantly with its use. For slice selection, a high sensitivity is required. This was achieved with a simple random forest classifier. As the image data is large, SVM is unable to converge. Also the features differ significantly between slices and the data is not linear.

Thresholding techniques are used to remove outliers in slice selection. DWT removes high frequency components for tissue-type segmentation. The contralateral sides are compared to find differences based on intensity thresholding. The intensity and features are symmetric across a patient image, which might not be true if a normal template been used. This methodology does not require a training set or a normal template thus making the segmentation process faster in terms of computation time and also more reliable [24, 25, 26]. Using a normal template can result in errors in identification as some normal variants may appear different in terms of intensity and be identified as tumors [7]. Also a training set may not incorporate all the different types of tumors that are being tested for. Many studies use spatial fuzzy c means clustering for segmentation [44, 48]. This method is not fully automated as the use of fuzzy gives a number of segmented images as per the number of clusters and the tumor segmented image must be chosen manually.

Using this three step process, false positive cases were eliminated, increasing the accuracy to 95%. This paper reports a higher accuracy than most studies with a larger dataset and the use of classification and segmentation jointly

[18, 22, 45, 44].

4. Conclusion

In past, many groups have attempted to automate the classification and segmentation of brain tumor using different MR image sequences. However, the reported results have restrictions in terms of images used and use of training sets or templates for segmentation. In this work, a combined model has been proposed for both classification and segmentation subsequently to obtain a higher accuracy. The data set is large, varied and the training and testing set do not overlap. The algorithm is comprehensive and effective within a short computation time. It is a three step process which involves identifying patients with tumor, then extracting the abnormal slices followed by segmentation of the tumor. The entire patient data set is used for classification treating each scan set as a single image comprising of 12 slices. Passing all slices considered as abnormal through segmentation allows for normal patients that have been misclassified to be correctly classified in the third step. Approximation components of the original images are mapped contralaterally for tumor segmentation which unlike previous studies does not require a normal template or a training set. The overall accuracy of the method proposed is 95% with 100% specificity and 90% sensitivity. Future work could include improving accuracy and segmenting the various cell aggregates that differ in composition, within the tumor.